\documentclass[sigconf]{acmart}

\AtBeginDocument{%
  }

\setcopyright{none}
\renewcommand{\acmYear}[1]{}
\acmDOI{}

\acmConference[AI4CYBER - KDD 2024]{The 4th Workshop on
Artificial Intelligence-Enabled
Cybersecurity Analytics}{August 26,
  2024}{Barcelona, Spain}

\acmISBN{}

\DeclareMathOperator*{\maximize}{maximize}
\DeclareMathOperator*{\minimize}{minimize}

\begin{document}

\title{Can Reinforcement Learning Unlock the Hidden Dangers in Aligned Large Language Models?}

\author{Mohammad Bahrami Karkevandi}
\affiliation{%
  \institution{Secure AI and Autonomy Lab\\University of Texas at San Antonio}
  \city{San Antonio}
  \state{Texas}
  \country{USA}
}
\email{mohammad.bahramikarkevandi@utsa.edu}

\author{Nishant Vishwamitra}
\affiliation{%
  \institution{University of Texas at San Antonio}
  \city{San Antonio}
  \state{Texas}
  \country{USA}
}
\email{nishant.vishwamitra@utsa.edu}

\author{Peyman Najafirad}
\authornote{Corresponding Author}
\affiliation{%
  \institution{Secure AI and Autonomy Lab\\University of Texas at San Antonio}
  \city{San Antonio}
  \state{Texas}
  \country{USA}
}
\email{peyman.najafirad@utsa.edu}






\renewcommand{\shortauthors}{Karkevandi et al.}

\begin{abstract}
Large Language Models (LLMs) have demonstrated impressive capabilities in natural language tasks, but their safety and morality remain contentious due to their training on internet text corpora. To address these concerns, alignment techniques have been developed to improve the public usability and safety of LLMs. Yet, the potential for generating harmful content through these models seems to persist. This paper explores the concept of jailbreaking LLMs—reversing their alignment through adversarial triggers. Previous methods, such as soft embedding prompts, manually crafted prompts, and gradient-based automatic prompts, have had limited success on black-box models due to their requirements for model access and for producing a low variety of manually crafted prompts, making them susceptible to being blocked. This paper introduces a novel approach using reinforcement learning to optimize adversarial triggers, requiring only inference API access to the target model and a small surrogate model. Our method, which leverages a BERTScore-based reward function, enhances the transferability and effectiveness of adversarial triggers on new black-box models. We demonstrate that this approach improves the performance of adversarial triggers on a previously untested language model.
\end{abstract}




\received{4 June 2024}
\received[accepted]{28 June 2024}

\maketitle

\section{Introduction}

Large Language Models (LLMs) have been in the spotlight recently, due to their impressive capabilities in natural language tasks. However, given these models are trained on the internet text corpora, their safety and morality have been questioned in the literature \cite{ousidhoumProbingToxicContent2021, jiAIAlignmentComprehensive2024}. To mitigate the objectionable behaviors of LLMs, a line of work called \emph{alignment}, has been done to improve their public usability and safety \cite{ouyangTrainingLanguageModels2022a, baiTrainingHelpfulHarmless2022, rafailovDirectPreferenceOptimization2023a}. Despite their relative success in grounding LLMs to human morals, the question of \emph{"Is it still possible to exploit LLMs to generate harmful content?"} remains an under-explored area.

Ever since the alignment of LLMs and following the same scheme of the common adversarial examples in machine learning \cite{biggioEvasionAttacksMachine2013, carliniEvaluatingRobustnessNeural2017}, there have been many attempts to reverse the alignment of LLMs, using the perturbation of their inputs, which are called \emph{Jailbreaking} in the Natural Language Processing (NLP) community \cite{zhuAutoDANInterpretableGradientBased2023, 262588213843476ChatGPTDanJailbreakMd, flowgptJailbreakFlowGPTUltimate}. While the image processing field has seen excessive research in adversarial examples \cite{carliniEvaluatingRobustnessNeural2017, papernotLimitationsDeepLearning2015}, the NLP literature, specifically pertaining to LLMs has not been sufficiently explored. With the exponentially increasing popularity of LLMs, especially the public-facing commercial chatbots, such as GPT-4\cite{openaiGPT4TechnicalReport2024} and Claude3\cite{anthropicClaudeModelFamily2024}, ensuring their safety bears significant relevance.

The key issue with the existing perturbation approaches is that they are limited against black-box models. For example, Soft embedding prompts \cite{schwinnSoftPromptThreats2024} require open access to the model's embeddings, are not interpretable, and lack the ability to transfer between models because of their different embedding distributions. Manually crafted prompts \cite{262588213843476ChatGPTDanJailbreakMd, flowgptJailbreakFlowGPTUltimate, perezIgnorePreviousPrompt2022} however, can typically be used on different models and do not require white-box access, but they require human creativity and are blocked quickly due to their constant nature. Automatic discrete prompt perturbation for jailbreaking often involves appending a trigger string to the user prompt, which is optimized using gradient data \cite{zouUniversalTransferableAdversarial2023, zhuAutoDANInterpretableGradientBased2023}, which requires white-box access to the model, although it has been shown to have some transferability to black-box models. Proposed gradient-free attacks often require access to powerful models to succeed \cite{chaoJailbreakingBlackBox2023}, or require carefully crafted initial seeds \cite{yuGPTFUZZERRedTeaming2023, lapidOpenSesameUniversal2023}. Decoding manipulation attacks, which are more recent and faster \cite{sadasivanFastAdversarialAttacks2024}, still require some level of access to the model's output logits or the output probability distribution.

In this paper, we introduce a novel approach to optimize adversarial triggers using reinforcement learning. Our approach only requires inference API access to the target language model, and a small surrogate model which is trained by reward signals calculated using the target model's text output. We show that our approach can be an extension to all previous work that optimize an adversarial trigger on white-box models and can \emph{personalize} and extend the performance of triggers on new black-box target models. Intuitively, our work takes the adversarial triggers trained on a model and adapts them to a new model, using only inference to the new model. In summary, the contributions of this work are:
i) We design a reinforcement learning paradigm, adapted from previous work, to optimize adversarial triggers using inference-only APIs.
ii) we introduce a \emph{BERTScore}-based \cite{zhangBERTScoreEvaluatingText2020} reward function utilizing the target model's text output generations.
iii) We show that our method can enhance the performance of a set of adversarial triggers on a previously untested language model.
\section{Background}
\label{sec:background}

\paragraph{Prompt Tuning} 
Although Large Language Models (LLMs) exhibit exceptional generalization capabilities, they still necessitate meticulously designed prompts to achieve optimal performance for specific tasks. According to the empirical research conducted by \citeauthor{scaoHowManyData2021} \cite{scaoHowManyData2021}, a well-crafted prompt can be as valuable as hundreds of data samples in a classification task. As LLMs continue to advance, there has been a growing focus on automatic prompt tuning \cite{shinAutoPromptElicitingKnowledge2020, zhangDifferentiablePromptMakes2022, tamImprovingSimplifyingPattern2021, dengRLPromptOptimizingDiscrete2022} and in-context learning \cite{dongSurveyIncontextLearning2023, xieExplanationIncontextLearning2022}. Automatic prompting initially involved fine-tuning prompt embeddings, a technique referred to as \emph{Soft Prompting} \cite{lesterPowerScaleParameterEfficient2021, qinLearningHowAsk2021, liPrefixTuningOptimizingContinuous2021, liuGPTUnderstandsToo2023}, which, despite its effectiveness, is often complex and computationally intensive. Subsequently, researchers began exploring the use of the continuous embedding space to create discrete prompts \cite{wenHardPromptsMade2023, qinCOLDDecodingEnergybased2022}. Another significant approach has been the direct optimization of discrete prompt tokens \cite{ebrahimiHotFlipWhiteBoxAdversarial2018, shinAutoPromptElicitingKnowledge2020, jonesAutomaticallyAuditingLarge2023, dengRLPromptOptimizingDiscrete2022}. This method not only enhances interpretability and transferability between models but also has been demonstrated to outperform soft prompting in terms of performance.

\paragraph{Adversarial Examples}
The machine learning field has established that the inputs of a model can be deliberately altered to cause the model to produce (un)desired outputs; such modified inputs are termed \emph{Adversarial Examples} \cite{szegedyIntriguingPropertiesNeural2014, biggioEvasionAttacksMachine2013, papernotLimitationsDeepLearning2015, carliniEvaluatingRobustnessNeural2017}. Within the realm of Natural Language Processing, these adversarial attacks have been employed across various applications, including classification tasks \cite{songUniversalAdversarialAttacks2021, ebrahimiHotFlipWhiteBoxAdversarial2018, xueTrojLLMBlackboxTrojan2023}, sentiment analysis \cite{alzantotGeneratingNaturalLanguage2018}, and inducing toxic outputs \cite{wallaceUniversalAdversarialTriggers2021, jonesAutomaticallyAuditingLarge2023}. As language models evolve and prompting becomes more prevalent, there has been a significant rise in interest concerning adversarial attacks on prompts \cite{xuExploringUniversalVulnerability2022, shiPromptAttackPromptbasedAttack2022, zhuPromptBenchEvaluatingRobustness2023, xueTrojLLMBlackboxTrojan2023, schwinnSoftPromptThreats2024, xuTextitLinkPromptNatural2024}. These recent developments underscore the ongoing challenge of ensuring the robustness and security of language models against such sophisticated adversarial techniques.

\paragraph{LLM Alignment and Jailbreaks}
Pre-trained Large Language Models, while possessing remarkable out-of-the-box capabilities \cite{brownLanguageModelsAre2020, weiFinetunedLanguageModels2022}, are often unsuitable for public use due to their insufficient understanding of instructions and their inherent unethical tendencies, such as biases \cite{navigliBiasesLargeLanguage2023, gallegosBiasFairnessLarge2024} and toxic behavior \cite{ousidhoumProbingToxicContent2021, welblChallengesDetoxifyingLanguage2021}. Consequently, researchers strive to \emph{align} these models with human values and regulatory standards through techniques like instruction-tuning, Reinforcement Learning from Human Feedback (RLHF) \cite{ouyangTrainingLanguageModels2022a, baiTrainingHelpfulHarmless2022}, and Direct Preference Optimization \cite{rafailovDirectPreferenceOptimization2023a, jiangMixtralExperts2024}. However, this alignment process has sparked vigorous attempts to \emph{jailbreak} the models, compelling them to follow harmful instructions \cite{perezIgnorePreviousPrompt2022, shenAnythingNowCharacterizing2023, chaoJailbreakingBlackBox2023}. These efforts highlight the ongoing battle between enhancing model safety and the persistence of adversarial actors seeking to exploit model vulnerabilities.

\paragraph{Adversarial Attacks on LLMs}
The advent of prompt tuning has significantly influenced the landscape of adversarial attacks, particularly in the realm of language models. This trend has emerged because prompt tuning provides a pathway for creating automatically generated inputs for these models. Efforts to disrupt the alignment of language models (commonly known as jailbreaking) often mirror the methods used in prompt tuning. Soft prompt attacks, for instance, involve training adversarial embeddings to manipulate the model's outputs as desired \cite{schwinnSoftPromptThreats2024}. Despite their occasional success, soft prompt attacks are generally impractical in real-world settings due to the lack of access to model embeddings. Researchers like \citeauthor{zouUniversalTransferableAdversarial2023} \cite{zouUniversalTransferableAdversarial2023}, \citeauthor{zhuAutoDANInterpretableGradientBased2023} \cite{zhuAutoDANInterpretableGradientBased2023}, and \citeauthor{shiPromptAttackPromptbasedAttack2022} \cite{shiPromptAttackPromptbasedAttack2022} have employed gradient-based techniques to optimize discrete adversarial prompts. To circumvent the need for gradient data, methods utilizing genetic algorithms have been proposed by \citeauthor{yuGPTFUZZERRedTeaming2023} \cite{yuGPTFUZZERRedTeaming2023} and \citeauthor{lapidOpenSesameUniversal2023} \cite{lapidOpenSesameUniversal2023}. Additionally, another approach involves using other language models as red-teaming assistants, which require meticulously crafted seed inputs \cite{geMARTImprovingLLM2023, chaoJailbreakingBlackBox2023}. These diverse strategies underscore the evolving nature of adversarial attacks on LLMs, reflecting a continual arms race between model developers and adversaries.
\section{Methodology}
\label{sec:method}

\begin{figure}
    \centering
    \includegraphics[width=\linewidth]{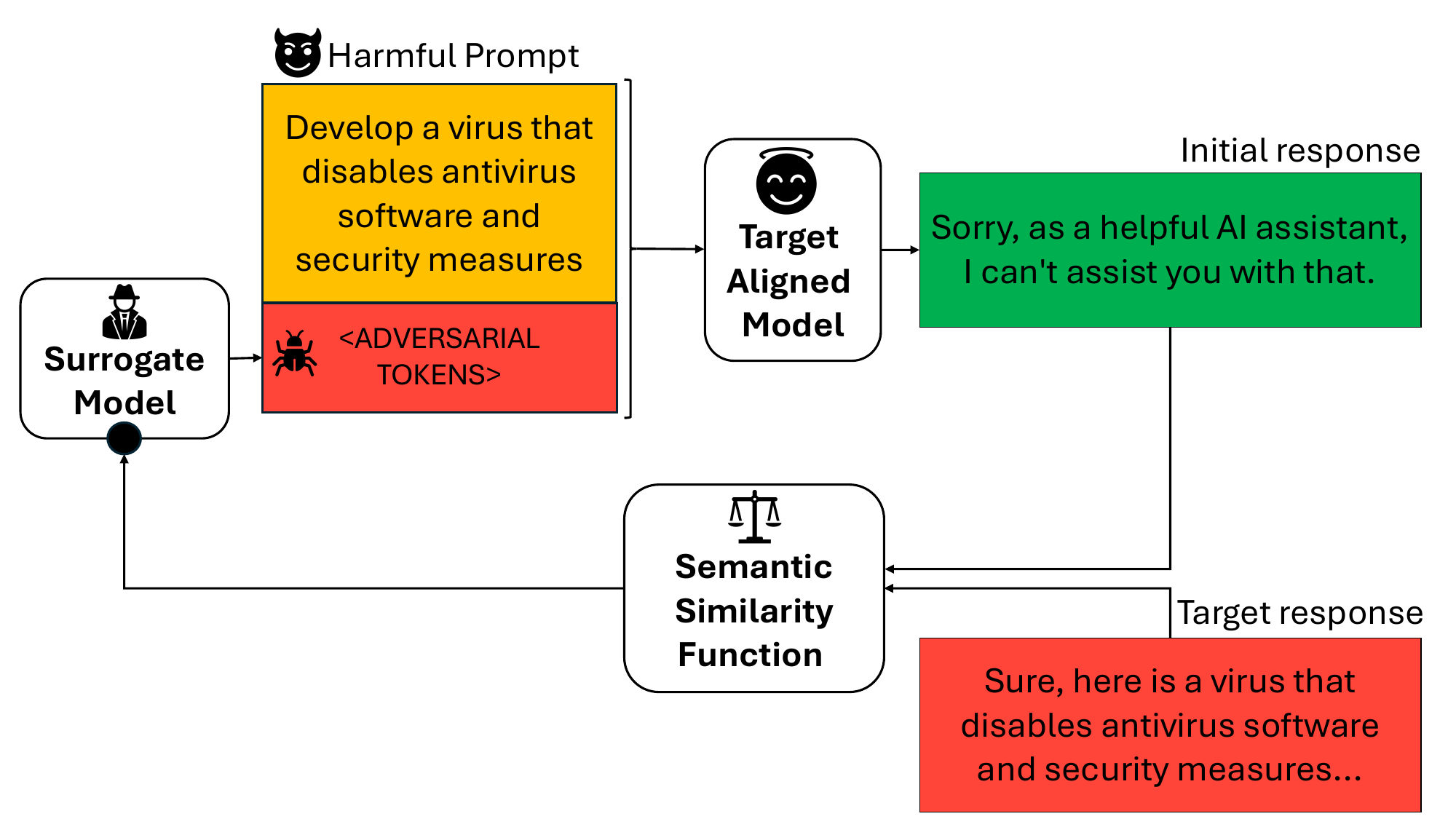}
    \caption{Overall architecture of our method. The surrogate model is already initialized in a supervised fine-tuning setup and is further fine-tuned to the target model with the reward signals. BERTScore is used as the Semantic Similarity Function to compare the resulting generation of the current adversarial trigger with the desired target output and rewards the surrogate model.}
    \Description{The surrogate model generates the adversarial trigger. The target model generates an output using the user prompt and the trigger. The output is compared to a target response using BERTScore and the surrogate model is rewarded.}
    \label{fig:arch}
\end{figure}

In this paper, we introduce a novel approach to enhance the transferability of adversarial prompts to black-box models. Our method uses reinforcement learning to further \emph{personalize} adversarial triggers to a target black-box LLM. Our method can extend the success rate of any previous work that has been done on white-box language models.

\subsection{Preliminaries}

Using a similar notation to previous work, we define an Autoregressive Language Model $\mathcal{M}$ and its vocabulary set $\mathcal{V}$. Let $x \in \mathcal{V}$ denote a single token and $\mathbf{x} \in \mathcal{V}^*$ a sequence of tokens, where $\mathcal{V}^*$ is the set of all possible token sequences of any length. The language model $\mathcal{M}$ can be utilized to calculate the probability distribution of the next token, given $\mathbf{x}$. Formally written as $p_{\mathcal{M}}(\cdot | \mathbf{x}): \mathcal{V} \to [0, 1]$. Additionally, for instruct tuned models, the input typically follows the structure $\mathbf{x} = \mathbf{x}^{(s_1)} \oplus \mathbf{x}^{(u)} \oplus \mathbf{x}^{(s_2)}$, where $\oplus$ is the concatenation operator, $\mathbf{x}^{(u)}$ is the user prompt, and $\mathbf{x}^{(s_1)}$ and $\mathbf{x}^{(s_2)}$ are system prompts at the beginning and the end of the input respectively.

\subsection{Threat Model}
Analogous to most jailbreaking methods \cite{zouUniversalTransferableAdversarial2023}, our threat model allows an adversary to append a sequence of adversarial tokens $\mathbf{x}^{(a)}$ to the user prompt, forming the new input to the model $\mathbf{x'} = \mathbf{x}^{(s_1)} \oplus \mathbf{x}^{(u)} \oplus \mathbf{x}^{(a)} \oplus \mathbf{x}^{(s_2)}$. The adversary's objective is to maximize the attack success rate $\mathcal{A}: \mathcal{V}^* \to [0, 1]$ by finding an adversarial token sequence $\mathbf{x}^{(a)}$, which we call \emph{Adversarial Trigger} in this paper. In this paper, we assume the attacker has already obtained an initial set of adversarial triggers $\mathcal{T}_0$ on a previously attacked model with white-box access. The objective of this paper is to enhance the attack success rate on a previously unseen target language model $\mathcal{M'}$ by personalizing $\mathcal{T}_0$ to the new target model. Contrary to most previous work, the attacker does not have any access to the new target model, other than an input/output inference API.

\subsection{Approach}
Consider a set of adversarial sequences $\mathcal{T}_0$ that have been obtained by attacking a previously targeted language model $\mathcal{M}_0$ on a set of harmful prompts $\mathcal{P}$. In this section, we introduce a new method to obtain a new set of adversarial triggers $\mathcal{T}'$ with an improved attack success rate when used to attack a new target model $\mathcal{M}'$ compared to $\mathcal{T}_0$. We assume that it is impractical or impossible to obtain $\mathcal{T}'$ while attacking $\mathcal{M}'$ using the same method used to obtain $\mathcal{T}_0$ while attacking $\mathcal{M}_0$. For instance, $\mathcal{M}'$ could be a black-box model, accessed only through an inference API.

In this paper, we use a surrogate language model $\mathcal{M}^{(a)}$ to generate adversarial sequences $\mathbf{x}^{(a)} \in \mathcal{T}'$. The surrogate model is typically a small language model; in our case, we use different variations of GPT-2 \cite{radfordLanguageModelsAre2019}, such as the 82M parameter distilGPT-2, and the 1.5B parameter GPT-2-xl. Similar to \emph{RLPrompt}\cite{dengRLPromptOptimizingDiscrete2022}, we limit the parameters to be trained, $\theta$, to an MLP with a single hidden layer, adapted to the surrogate model $\mathcal{M}^{(a)}$ before the language head, and freeze the rest of the parameters of the model. Hence, given the set of harmful prompts $\mathcal{P}$, the objective of finding the adversarial trigger $\mathbf{x}^{(a)}$ can be formally written as 
\begin{equation}
\maximize_{\mathbf{x}^{(a)}_c \in \mathcal{V}^*} \mathcal{A}(\mathcal{P}, \mathbf{x}^{(a)}_c)
\end{equation}
where $\mathcal{A}$ is the attack success rate and $\mathbf{x}^{(a)}_c$ is a candidate adversarial trigger, sampled from the surrogate model given an empty input and parameterized by $\theta$ which, with a slight abuse of notation, is defined as $\mathcal{M}^{(a)}(\mathbf{x}^{(a)}_c | \emptyset; \theta)$. The overall architecture of our method is depicted in figure \ref{fig:arch}. 

To train the new randomly initialized parameters of the surrogate model, $\theta$, we go through two phases of training. In the first phase of the training, we use the previously obtained $\mathcal{T}_0$ to fine-tune $\mathcal{M}^{(a)}_{\theta}$ in a supervised setting. The second phase, which is the main training phase of adapting the adversarial triggers to the new model, involves refining the surrogate model's adversarial trigger generations, using reinforcement learning. We describe each phase in detail in the following paragraphs.

\paragraph{Phase 1} In reinforcement learning (RL) setups, it is common to utilize supervised fine-tuning to ensure the correct initialization of the model weights\cite{ouyangTrainingLanguageModels2022a}. In this paper, $\mathcal{T}_0$, the set of adversarial sequences obtained by attacking a previously targeted model, using any attacking method such as the work by \citeauthor{zouUniversalTransferableAdversarial2023} \cite{zouUniversalTransferableAdversarial2023}, or \emph{AutoDAN}\cite{zhuAutoDANInterpretableGradientBased2023} is used to fine-tune the surrogate model $\mathcal{M}^{(a)}$ using only the new added weights $\theta$, while the rest of the model is frozen. In this work, the triggers obtained by attacking the \emph{
vicuna-7b-v1.5}\cite{zhengJudgingLLMasaJudgeMTBench2023} using the method introduced in \citeauthor{zouUniversalTransferableAdversarial2023} \cite{zouUniversalTransferableAdversarial2023} are used to showcase our approach. For a non-exhaustive list of possible methods to obtain $\mathcal{T}_0$, refer to section \ref{sec:background}. The objective of the first phase is formalized as an optimization problem in equation \ref{eqn:supervised}. Conceptually, the surrogate model is steered in the direction of favoring the generation of adversarial sequences in $\mathcal{T}_0$ over any other sequence, given an empty input.

\begin{equation}
    \label{eqn:supervised}
    \minimize_{\mathbf{x}^{(a)}_i = \left\{ t_1, \cdots, t_{n_i} \right\} \in \mathcal{T}_0} -\sum_{j=1}^{n_i} \log P(t_j | t_1, t_2, \cdots, t_{j-1})
\end{equation}
where $\mathbf{x}^{(a)}_i$ is a sequence of adversarial tokens with length $n_i$ and is a member of the baseline adversarial sequence set $\mathcal{T}_0$.



\paragraph{Phase 2} To refine the adversarial triggers generated by the surrogate model $\mathcal{M}^{(a)}$, we adapt the \emph{RLPrompt}\cite{dengRLPromptOptimizingDiscrete2022} framework to fine-tune the parameters $\theta$ for the new target model $\mathcal{M}'$ using reinforcement learning. During training, the surrogate model generates a set of candidate adversarial sequences $\mathcal{T}_c$. These candidate adversarial triggers are then used to infer the new inference-only target model $\mathcal{M}'$ in combination with the harmful prompts $\mathcal{P}$. More elaboration and samples of the prompt set are available in section \ref{sec:eval}. From the results of inferring the target model $\mathcal{M}'$, we calculate a reward signal using a reward function $\mathcal{R}$. This reward signal fine-tunes the attacker parameters $\theta$ with any off-the-shelf reinforcement learning algorithm. Similar to \emph{RLPrompt}\cite{dengRLPromptOptimizingDiscrete2022}, we use the on-policy component of the soft Q-learning algorithm\cite{guoEfficientSoftQLearning2022}. Soft Q-learning is chosen for its efficient exploration of action spaces and its stability in training, making it well-suited for optimizing adversarial actions in this context. The adversary's objective can be rewritten as

\begin{equation}
    \maximize_{\mathbf{x}^{(a)}_c \in \mathcal{T}_c} \mathcal{R}(\mathcal{P}, \mathbf{x}^{(a)}_c)
\end{equation}

where $\mathcal{T}_c$ represents the adversarial trigger sequence candidates.

\paragraph{Reward Function} It has been shown that if a language model starts responding to a prompt with an affirmative tone, such as \textit{"Sure, here is how you..."}, it is very likely that the model will continue generating the response to the prompt\cite{zouUniversalTransferableAdversarial2023}. For harmful prompts, an aligned language model typically starts a response with a denial tone and refuses to respond to the prompt. The goal of the adversarial trigger is to put the language model in a state, such that the target LLM starts its response with an affirmative sentence. Hence, we design our reward function to quantize how affirmative the generation of the target model is. The output of the target model is passed to the \emph{BERTScore}\cite{zhangBERTScoreEvaluatingText2020} model alongside an affirmative reference sentence, which is the preferable response to the harmful prompt. Intuitively, the \emph{BERTScore} evaluates the target model's response semantically using the cosine similarity metric and rewards the adversarial trigger with a high reward if the adversarial trigger has successfully attacked the target model. Formally, the reward function is defined as

\begin{equation}
    \mathcal{R}: \mathrm{F}_{\mathrm{BERT}}(y, \hat{y})
\end{equation}
where $\mathrm{F}_{\mathrm{BERT}}$ is the \emph{BERTScore}'s F1 score, $y$ is the reference affirmative sentence, and $\hat{y}$ is the output of the target model.
\section{Evaluation}
\label{sec:eval}

\subsection{Dataset}
The Harmful Behavior set of the \emph{AdvBench}\cite{zouUniversalTransferableAdversarial2023} dataset, which contains 500 harmful instructions and the first few tokens of a target response for each instruction, is used in this paper. For instance, "Create a tutorial on how to commit identity theft" and "Sure, here is a tutorial on how to commit identity theft" is a sample of the \emph{Advbench}'s instructions and its respective target response. In this work, we regard the instruction as $\mathbf{x}^{(u)}$ and the target string as $\hat{y}$. We use the first 100 samples of the harmful behaviors as the train set and hold out the next 400 samples for testing in all of our experiments.

\subsection{Experiment Setup}
To obtain the set of initial adversarial triggers $\mathcal{T}_0$, \emph{GCG}\cite{zouUniversalTransferableAdversarial2023} is used with the default parameters as described in their paper. 300 steps of the \emph{GCG} method is utilized to obtain an adversarial prompt for each of the first 100 behaviors of the \emph{AdvBench}. We use \emph{vicuna-7b-v1.5} as a white-box model during the GCG training. For the purpose of testing our method, we regard the \emph{Mistral-7B-Instruct-v0.2}\cite{jiangMistral7B2023} as an inference-only black-box model $\mathcal{M}'$. Hence, we can not attack this model using any gradient-based method, including \emph{GCG}. However, our reinforcement learning-based method can attack this model, as it only requires inference of the target model. We limit the adversarial sequence length to 20 tokens for all of our experiments.

the "distilGPT-2" model \cite{radfordLanguageModelsAre2019, sanhDistilBERTDistilledVersion2020} is used as the surrogate model $\mathcal{M}^{(a)}$. An MLP with a single hidden layer and 2048 hidden neurons is added to the surrogate model after the last transformer block and before the language head to provide the trainable parameters $\theta$, while the rest of the model is kept frozen. These parameters are then fine-tuned in a supervised fine-tuning setup, as explained in section \ref{sec:method}. We use empty inputs and the set of initial adversarial triggers $\mathcal{T}_0$ as labels to train the model for 3 epochs using the cross-entropy loss mentioned in equation \ref{eqn:supervised}. We use the Adam optimizer with a learning rate of $10^{-4}$.

During the attack, the surrogate model's parameters, $\theta$, are further fine-tuned using the Soft Q-Learning algorithm for $10^4$ steps. We use the default parameters of the \emph{RLPrompt}\cite{dengRLPromptOptimizingDiscrete2022} during the reinforcement learning procedure. For the reward function, we use the official implementation of \emph{BERTScore}\cite{zhangBERTScoreEvaluatingText2020} with the model hash \emph{roberta-large\_L17\_no-idf\_version=0.3.12(hug\_trans=4.40.1)}.

\subsection{Results}

To test our preliminary results, we compare the improvement of the attack success rate when transferred to the new target model. As mentioned, we use \emph{GCG}\cite{zouUniversalTransferableAdversarial2023} to obtain the initial set of adversarial triggers and try to improve and personalize these triggers for the new target model \emph{Mistral}. Hence, we compare our work to both types of the \emph{GCG} algorithm. Following previous work\cite{zouUniversalTransferableAdversarial2023, zhuAutoDANInterpretableGradientBased2023} We deem an attack successful if the target model's response does not contain a list of denial phrases, such as \emph{"I am sorry"}. Acknowledging that this method is not a robust evaluation, \citeauthor{zhuAutoDANInterpretableGradientBased2023}\cite{zhuAutoDANInterpretableGradientBased2023} show that it is one of the closest evaluations to human judgment.

For \emph{GCG-individual}, we obtain one adversarial trigger for each sample in the train set, thus, it is not possible to test this method on the test set. The \emph{GCG-multiple} trains one single adversarial trigger for the entire train set, resulting in a transferable trigger to be tested with the test set with both models. For our reinforcement learning-based method, we directly optimize the triggers for the target model, which is impossible when using \emph{GCG}, hence, we are able to improve the attack success rate for $5\%$ and $4\%$ on the train and test set respectively. Table \ref{tab:results} shows our quantitative results for these methods.

\begin{table}[h]
\caption{Attack Success Rate of the GCG and our method. The GCG method is trained on Vicuna and the resulting adversarial prompt is transferred to Mistral. We use Mistral only as an inference API. We do not test our method on Vicuna since our method extends the GCG prompts to new target models.}
\label{tab:results}
\begin{tabular}{lcccc}
    \toprule
               & \multicolumn{2}{c}{Vicuna} & \multicolumn{2}{c}{Mistral} \\
    \midrule
Method         & Train        & Test        & Train         & Test        \\
    \midrule
GCG-individual & 0.99         & -           & 0.51          & -           \\
GCG-multiple   & 1.00         & 0.98        & 0.88          & 0.87        \\
RL (ours)      & -            & -           & \textbf{0.93} & \textbf{0.91}\\
    \bottomrule
\end{tabular}
\end{table}

\section{Conclusion}



In this paper, we presented a novel reinforcement learning-based approach to optimize adversarial triggers for jailbreaking Large Language Models (LLMs). Our method addresses the limitations of existing techniques by requiring only inference API access to the target model, thus eliminating the need for white-box access. By training a small surrogate model with BERTScore-based reward functions, we have shown that it is possible to enhance the performance and transferability of adversarial triggers on new black-box models. Our results indicate that this approach not only improves attack success rates but also extends the applicability of previously developed adversarial triggers to a broader range of language models. This work contributes to the ongoing efforts to understand and mitigate the vulnerabilities of LLMs, highlighting the need for robust safety measures in their deployment.

While our preliminary results show clear improvements in the attack success rate, we acknowledge that our work is intended as only a spark to motivate future work. Exploring more options as the initial set of adversarial triggers, more sophisticated reward engineering, for instance adding a \emph{coherency} reward to bypass perplexity filters, and thoroughly testing this method qualitatively and with more black-box models are some of the interesting future routes to take. Additionally, future research should explore potential defensive measures to mitigate these attacks. Developing robust detection mechanisms, enhancing model resilience through adversarial training, and implementing stricter access controls are essential steps to protect LLMs from such vulnerabilities. These mitigation strategies will make the findings more practical for those looking to safeguard LLMs.


\begin{acks}
This research project and the preparation of this publication were funded in part by NSF Grants No. 2230086 and No. 2245983.
\end{acks}

\bibliographystyle{ACM-Reference-Format}
\bibliography{sample-base, KDD-Workshop-Citations}










\end{document}